\documentclass[sigconf]{acmart}
\AtBeginDocument{%
  }

\begin{document}

\title{Metadata Extraction Leveraging Large Language Models}

\author{Cuize Han}
\authornote{Both authors contributed equally to this research.}
\email{chan@box.com}
\orcid{0000-0003-3957-0687}
\author{Sesh Jalagam}
\authornotemark[1]
\email{sjalagam@box.com}
\affiliation{%
  \institution{Box AI Platform}
  \city{Redwood City}
  \state{California}
  \country{USA}
}









\begin{abstract}
The advent of Large Language Models has revolutionized tasks across domains, including the automation of legal document analysis—a critical component of modern contract management systems. This paper presents a comprehensive implementation of LLM-enhanced metadata extraction for contract review, focusing on the automatic detection and annotation of salient legal clauses. Leveraging both the publicly available Contract Understanding Atticus Dataset (CUAD) and proprietary contract datasets, our work demonstrates the integration of advanced LLM methodologies with practical applications. We identify three pivotal elements for optimizing metadata extraction: robust text conversion, strategic chunk selection, and advanced LLM-specific techniques, including Chain of Thought (CoT) prompting and structured tool calling. The results from our experiments highlight the substantial improvements in clause identification accuracy and efficiency. Our approach shows promise in reducing the time and cost associated with contract review while maintaining high accuracy in legal clause identification. The results suggest that carefully optimized LLM systems could serve as valuable tools for legal professionals, potentially increasing access to efficient contract review services for organizations of all sizes.
\end{abstract}

\ccsdesc[500]{Information systems~Retrieval models and ranking}
\ccsdesc[500]{Computing methodologies~Information extraction}
\ccsdesc[500]{Applied computing~Law, social and behavioral sciences}

\keywords{Metadata Extraction, LLM, Chunk re-ranking, OCR, Chain of Thought}



\maketitle

\section{Introduction}



\subsection{Metadata Extraction}
Metadata extraction is the automated process of identifying and extracting structured information from unstructured or semi-structured documents. In today's digital age, this task has become increasingly critical, particularly in legal and enterprise applications. Automated clause extraction and analysis from legal contracts has become increasingly critical in modern legal practice, where law firms spend approximately 50\% of their time on contract review \cite{hendrycks2021cuad}. This task involves identifying and extracting structured information such as governing law clauses, termination provisions, and liability terms from complex legal documents. With billing rates for lawyers at large law firms typically ranging from \$500-\$900 per hour in the US \cite{hendrycks2021cuad}, manual contract review creates substantial costs for organizations. Furthermore, the prohibitive cost of legal review means many small businesses and individuals often sign contracts without thorough analysis, potentially exposing themselves to unfavorable terms \cite{upcounselURL}. In the legal domain, automatic contract analysis requires precise extraction of key elements such as parties involved, effective dates, termination clauses, and financial terms \cite{sleimi2018automated}. This automation not only accelerates contract review processes but also reduces human error and enables large-scale contract analytics \cite{sleimi2021automated}. 

Beyond legal applications, metadata extraction serves as a fundamental building block for various enterprise document management tasks. For instance, organizations can automatically classify documents based on their security levels by extracting and analyzing specific metadata fields \cite{sajid2023novel}. This capability is crucial for maintaining compliance with data protection regulations and implementing proper access controls. Furthermore, metadata extraction facilitates efficient document retrieval, content organization, and workflow automation across different business units \cite{haynes2004metadata}.

\subsection{Large Language Models}
Large Language Models (LLMs) represent a significant advancement in natural language processing, characterized by their ability to understand and generate human-like text through massive-scale pre-training on diverse datasets \cite{minaee2024large}. These models, such as GPT-4, Claude, and Gemini, have demonstrated remarkable capabilities in various language understanding and generation tasks \cite{chang2024survey}.

For metadata extraction, LLMs offer unique advantages over traditional rule-based or specialized machine learning approaches. Their deep contextual understanding enables them to accurately identify relevant information even when it appears in varying formats or requires complex reasoning. Importantly, our approach does not involve fine-tuning LLMs specifically for metadata extraction tasks. Instead, we utilize LLMs as the final component in our pipeline, where they analyze preprocessed text chunks and generate structured JSON outputs containing the extracted metadata.
This architecture offers several key benefits:
\begin{itemize}
\item Flexibility to experiment with and switch between different LLM providers.
\item Ability to leverage continuous improvements in base models without retraining.
\item Reduced dependency on task-specific training data.
\item Enhanced maintainability and adaptability of the system.
\end{itemize}

\subsection{Related Works}

Previous research in metadata extraction can be broadly categorized into several approaches:

Traditional Rule-based Systems: Early works relied heavily on regular expressions and predefined patterns \cite{tang2010regular}, which proved brittle when faced with document variations.

Machine Learning Approaches: Researchers have explored various ML techniques including:

Conditional Random Fields (CRF) for sequence labeling \cite{souza2014arctic}, 

Deep learning models specifically trained for extraction tasks \cite{liu2018automatic},

Transformer-based approaches with fine-tuning \cite{la2022transformer}.

LLM Integration: Recent work has begun exploring the use of LLMs for extraction \cite{savelka2023unlocking}.

Our Contributions:
Based on our research and experimentation, we make the following key contributions:

A comprehensive framework for metadata extraction that effectively combines text preprocessing, chunk selection, and LLM-based extraction
Novel techniques for optimal chunk selection that maximize extraction accuracy while minimizing LLM token usage
Empirical evaluation of various LLM prompting strategies, including Chain of Thought (CoT) and structured output generation
Extensive experimentation on both proprietary and public datasets, demonstrating the effectiveness and generalizability of our approach
Practical insights into implementing and scaling metadata extraction systems in enterprise environments

\section{Metadata Extraction Workflow}



\begin{figure}[h]
  \centering
  \includegraphics[width=\linewidth]{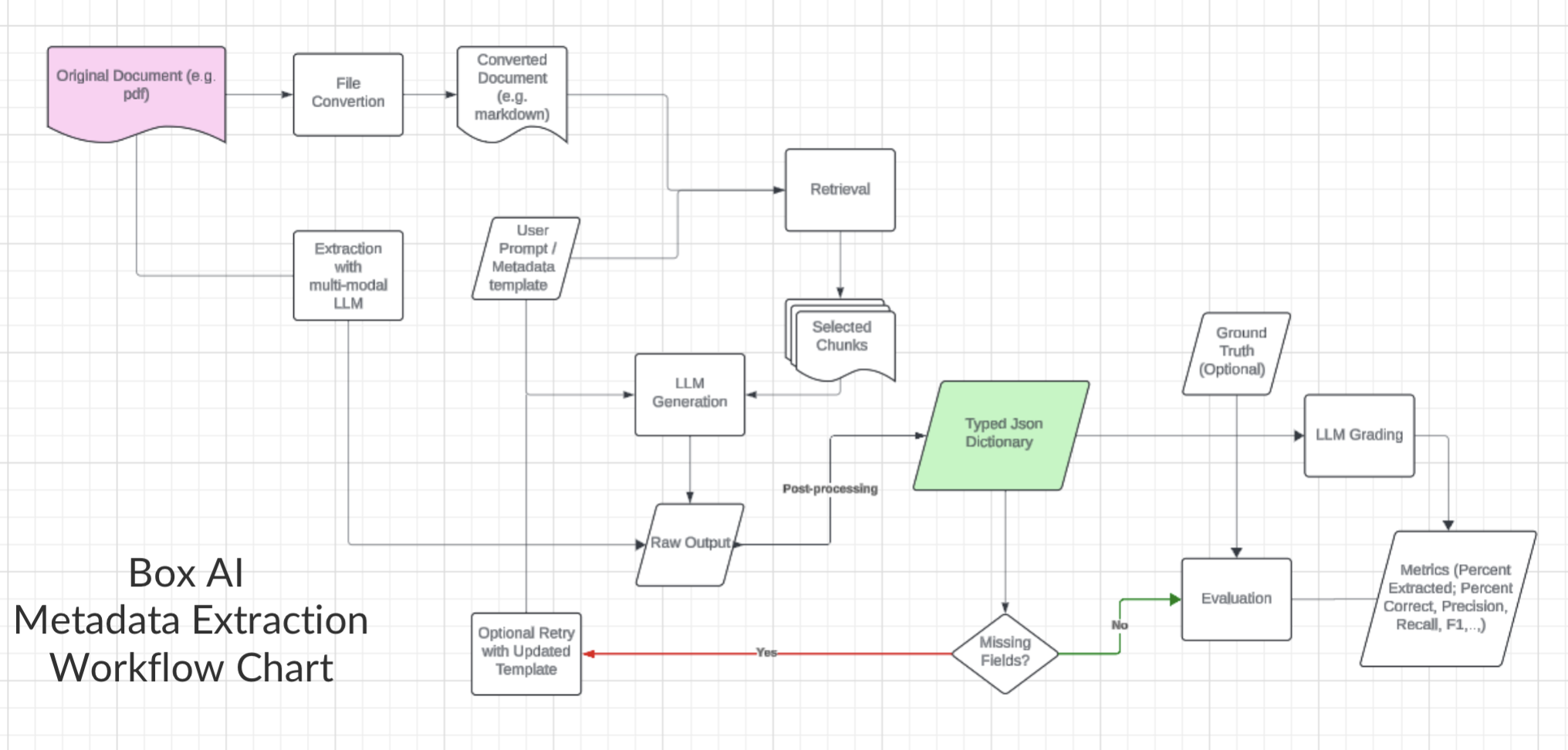}
  \caption{Simplified workflow of metadata extraction}
  \Description{as of 09/01/2024. including text conversion, chunk selection, prompting, offline evaluation, online grading, retrying, multi-modal extraction, automatic template enhancement, etc.}
  \label{system structure}
\end{figure}

Figure~\ref{system structure} presents a comprehensive overview of our metadata extraction system. The pipeline begins with an input document (typically in PDF format) and produces structured metadata output in JSON format, with optional quality assessment through LLM-based grading. Through extensive experimentation and deployment experience, we have identified three critical components that significantly impact both the quality and operational costs of metadata extraction: (1) text conversion and OCR preprocessing, (2) strategic chunk selection, and (3) LLM-based summarization and structuring.

While emerging multi-modal LLMs offer the possibility of direct metadata extraction from document images, our experiments indicate that this approach currently incurs prohibitively high costs for production deployment. Instead, our system employs a more efficient pipeline approach, with each component optimized for its specific task. The system also incorporates an innovative online grading mechanism using LLMs to assess extraction quality, enabling continuous monitoring and quality assurance.

In the following subsections, we examine each key component in detail, discussing their implementation, challenges, and our solutions for optimizing their performance. We also present empirical evidence demonstrating how these components collectively contribute to achieving high-quality metadata extraction while maintaining practical operational costs.

\subsection{Text conversion and OCR}

Text conversion and Optical Character Recognition (OCR) serve as the crucial first step in our metadata extraction pipeline. This component transforms various document formats, particularly PDFs, into machine-readable text while preserving essential structural information such as layout and formatting. The quality of this conversion directly impacts all subsequent steps in the pipeline, making it a fundamental determinant of the overall extraction accuracy.

For metadata extraction tasks, high-quality text conversion is particularly important for several reasons. First, accurate text extraction ensures that key information remains intact and properly sequenced, which is essential for LLMs to understand the document context correctly. Second, proper handling of document structure helps maintain the relationship between different metadata elements, such as associating values with their corresponding fields. Third, poor text conversion can introduce artifacts or errors that may confuse LLMs or lead to incorrect metadata extraction.
We conducted a comprehensive evaluation of various text conversion and OCR solutions, including both open-source and commercial options:

\begin{itemize}
    \item Evisort \cite{evisortURL}
    \item Azure Document Intelligence (azure-md) \cite{azureURL}
    \item Unstructured-IO (unstructured-md) \cite{unstructuredURL}
    \item Apryse OCR
    \item Apache Tika
    \item pdftotext
\end{itemize}

Our evaluation maintained a consistent LLM summarization configuration across all conversion methods to isolate the impact of text conversion quality. Based on extensive testing with our internal dataset, we selected Azure Document Intelligence as our primary conversion solution. This choice was driven by its optimal balance between conversion quality and operational costs. The detailed comparative analysis and performance metrics of different conversion methods are presented in section~\ref{ocr experiments}  of our experimental results.



\subsection{Chunk Selection}



Chunk selection is the process of identifying and extracting relevant portions of text from a document for metadata extraction. Despite the continuing expansion of LLMs' context windows, strategic chunk selection remains crucial for both improving extraction quality and optimizing operational costs. This component acts as an intelligent filter, ensuring that only the most relevant document segments are processed by the LLMs.

The importance of effective chunk selection extends beyond mere cost efficiency. Our experiments on the Contract Understanding Atticus Dataset (CUAD) demonstrate that smaller, well-selected context windows can actually outperform larger context windows using the same LLM. This counter-intuitive finding suggests that providing LLMs with more focused, relevant context can lead to more accurate metadata extraction than processing larger portions of text. The detailed results of these experiments are presented in section~\ref{chunk selection exp} .

Traditional chunk selection approaches, such as dense retrieval and cosine similarity-based methods, typically treat the document as a generic text corpus without considering the unique characteristics of metadata extraction tasks. These methods fail to leverage crucial structural information, including metadata field keys, descriptions, and expected value types. To address these limitations, we developed two novel techniques:

\begin{itemize}
    \item{\textbf{NER-enhanced Boosting}}: This technique incorporates Named Entity Recognition to identify and prioritize chunks containing entities relevant to the target metadata fields.
    \item{\textbf{Model-based Chunk Re-ranking}}: We implement a specialized re-ranking mechanism that considers both the structural aspects of the document and the specific requirements of each metadata field type.
\end{itemize}

\subsubsection{NER and Borda Re-ranking}

A key challenge in metadata extraction is ensuring that chunks containing essential information for each field are selected, particularly when dealing with large documents and complex metadata templates. We present NER-enhanced Boosting with Borda Re-ranking \cite{saari1985optimal}, a novel approach that combines named entity recognition with sophisticated scoring mechanisms to improve chunk selection quality.
The input for structured metadata extraction consists of a template defining multiple fields, each specified by a dictionary containing the field's key name, description, and optional type information (e.g., integer, string, enum, multiSelect). We augment this template by mapping fields to relevant entity labels from SpaCy's 18 pre-defined classes (e.g., DATE, EVENT, LAW). This mapping is determined through cosine similarity between field descriptions and entity label definitions.
For each chunk in the document, we compute four distinct scores:

\begin{itemize}
    \item{\textbf{Per-field Score}}: Cosine similarity between individual field information and chunk text
    \item{\textbf{Total-field Score}}: Cosine similarity between the aggregated template information and chunk text
    \item{\textbf{Per-field NER Score}}: Normalized count of named entities in the chunk matching the field's assigned entity labels
    \item{\textbf{Total-field NER Score}}: Normalized count of all relevant named entities in the chunk
\end{itemize}

These scores are combined using Borda re-ranking with tunable weights to create a final ranking. Our algorithm ensures coverage by maintaining a minimum percentage threshold of important chunks per field within the selected context window. This approach addresses a critical limitation of traditional cosine similarity-based selection, which can overlook chunks containing crucial field-specific information, especially in larger documents with numerous metadata fields.
One concrete template example:

\begin{verbatim}
"{
  "fields": [
    {
      "key": "Parties",
      "prompt": "List the parties who signed the contract.
      This usually contains names of organizations or individuals.",
      "type": "array",
      "ner_labels": [
        "PERSON",
        "ORG"
      ]
    },
    {
      "key": "Effective Date",
      "prompt": "Specify the date when the contract becomes
      effective.",
      "type": "date",
      "ner_labels": [
        "DATE"
      ]
    },
    {
      "key": "Most Favored Nation",
      "prompt": "Is there a clause ensuring the parties
      get the best terms offered to any third party?",
      "type": “enum”,
      "options": [
        {
          "key": "Yes"
        },
        {
          "key": "No"
        }
      ]
    } ] }"
\end{verbatim}

For fields like "Parties," our method ensures chunks containing relevant organization or person entities receive boosted scores, increasing their likelihood of selection. Similarly, chunks containing date entities are prioritized for date-related fields like "Effective Date."
The combination of NER-enhanced boosting and Borda re-ranking significantly improves both precision and recall of chunk selection, leading to better extraction quality in our experiments. Detailed performance comparisons with baseline approaches are presented in section~\ref{chunk selection exp}.

\subsubsection{Model-based Chunk Re-ranking}

\begin{figure}[h]
  \centering
  \includegraphics[width=\linewidth]{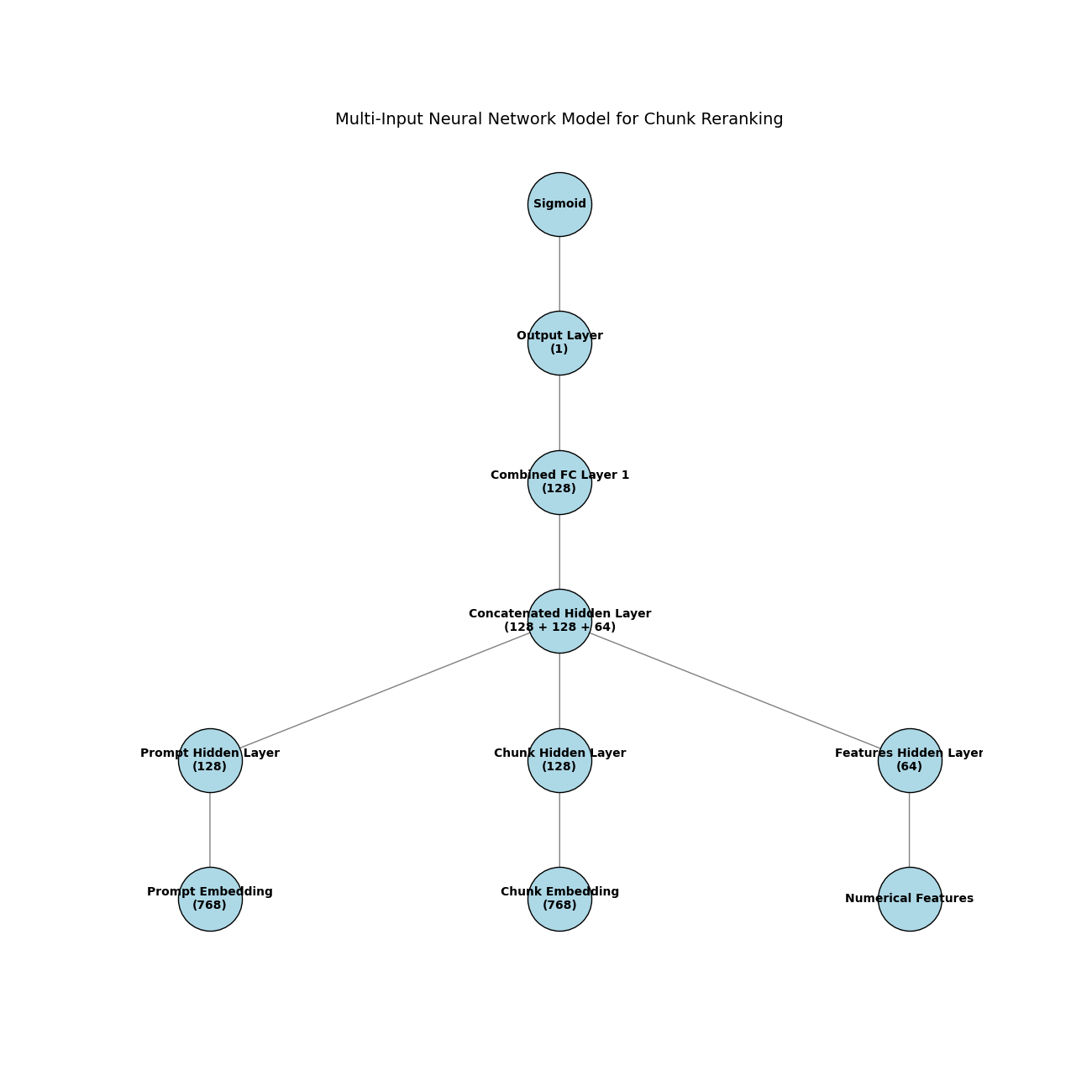}
  \caption{Neural network re-ranking model}
  \Description{nn chunk reranking model with mixed input structure }
  \label{nn_structure}
\end{figure}

Building upon the NER-enhanced Borda re-ranking approach, we developed a more sophisticated model-based chunk re-ranker that learns to predict the relevance of chunks for specific metadata fields. This approach offers two key advantages: (1) the ability to incorporate richer features through automatic feature engineering, and (2) the capacity to learn complex weight structures that surpass the linear combinations used in Borda voting.

Our model takes three inputs: the complete metadata template prompt, a specific field from the template, and a chunk of text. These inputs are featurized into numerical vectors that capture various aspects of their relationships and content. To maintain practical utility in production environments, we designed the model to be lightweight, minimizing both computational cost and latency.

To develop and validate this approach, we first established a performance ceiling through oracle experiments on the CUAD dataset. These experiments demonstrated that perfect chunk ranking could achieve an F1 score of 0.94, indicating significant potential for improvement over baseline methods. Guided by these results, we experimented with several model architectures, ultimately developing a neural network classifier that directly processes embeddings and additional features, see figure~\ref{nn_structure}.

The model incorporates multiple feature types:

\begin{itemize}
    \item Embedding-based features: Cosine similarities between chunk and field representations
    \item Text-based features: BM25 similarity scores
    \item Linguistic features: NER and POS tag-based features
    \item Structural features: Chunk position and length information
\end{itemize}

This model-based approach achieved significant improvements over baseline methods on CUAD dataset:
\begin{itemize}
    \item With an 8192-token context window: F1 score of 0.80 (vs. 0.75 baseline)
    \item With a 4096-token context window: F1 score of 0.76 (vs. 0.66 baseline)
\end{itemize}

The performance gains were particularly pronounced with smaller context windows, highlighting the model's efficiency in identifying the most relevant chunks. Detailed experimental results and ablation studies are presented in section~\ref{chunk selection exp}.

\subsection{LLM-based Information Synthesis}

\subsubsection{Chain of Thought Prompting}
The effectiveness of LLMs in metadata extraction heavily depends on prompt engineering. Beyond basic prompting guidelines such as clear instructions and consistent formatting, we found that Chain of Thought (CoT) prompting significantly improves extraction accuracy, particularly for fields requiring logical reasoning or complex calculations.

CoT prompting instructs the LLM to break down its reasoning process into explicit steps before providing the final answer. This step-by-step reasoning proves particularly valuable for:

\begin{itemize}
    \item Date calculations involving fiscal years or complex terms
    \item Financial calculations (e.g., payment schedules with compound terms)
    \item Conditional clauses where multiple sections need to be considered together
    \item Cross-referencing between different parts of the document and fields
\end{itemize}

While CoT prompting increases token usage and processing time, its benefits extend beyond improved accuracy. The detailed reasoning chains have helped us identify and correct annotation errors in both the CUAD dataset and our internal dataset, leading to better quality data.

\subsubsection{Structured Output through Tool Calling}

Tool calling (or function calling)~\cite{shen2024llm} capabilities of modern LLMs provide a robust mechanism for ensuring structured output format. By defining a JSON schema that specifies the expected structure and type constraints for each metadata field, we leverage the LLM's tool calling interface to enforce consistent output formatting.
Example tool definition:
\begin{verbatim}
{
  "name": "extract_metadata",
  "parameters": {
    "type": "object",
    "properties": {
      "document_name": {"type": "string"},
      "effective_date": {"type": "string", "format": "date"},
      "parties": {
        "type": "array",
        "items": {"type": "string"}
      }
    },
    "required": ["document_name", "effective_date", "parties"]
  }
}
\end{verbatim}

This approach offers several advantages including reduced format errors in LLM outputs, type validation at extraction time, improved extraction accuracy through structured constraints and better handling of null or missing values. Our experiments show that tool calling not only improves output consistency but also enhances overall extraction quality, likely because the structured format helps guide the LLM's reasoning process.
\section{Experiments}


Our primary evaluation metric is the average F1 score across all documents and fields. For each document and field, we calculate precision as \# correct extracted values / \# extracted values, recall as \# correct extracted values / \# correct values and define the F1 score as usual: 2 * (precision * recall) / (precision + recall). It handles cases where either true values or extracted values are arrays.

\subsection{Datasets}

We evaluate our approach using two distinct datasets. The first is the Contract Understanding Atticus Dataset (CUAD)\cite{hendrycks2021cuad}, a public benchmark dataset for contract analysis that encompasses a wide variety of contract types and metadata fields. It consists of over 500 contracts, each carefully labeled by legal experts to identify 41 different types of important clauses, for a total of more than 13,000 annotations. We focused on the 8 categories that are more aligned with the metadata extraction task including Document Name, Parties, Agreement Date, Effective Date, Expriation Date, Notice to Terminate Renewal and Governing Law. We used CUAD both for training our reranking models and for evaluation purposes. To assess the generalization capability of our approach, we also employ an internal dataset consisting of lease contracts with more than 40 fields for extraction, which provides complementary evaluation scenarios and helps validate the robustness of our methods across different document types.
\subsection{OCR and Text Conversion}

\label{ocr experiments}


\begin{figure}[h]
  \centering
  \includegraphics[width=\linewidth]{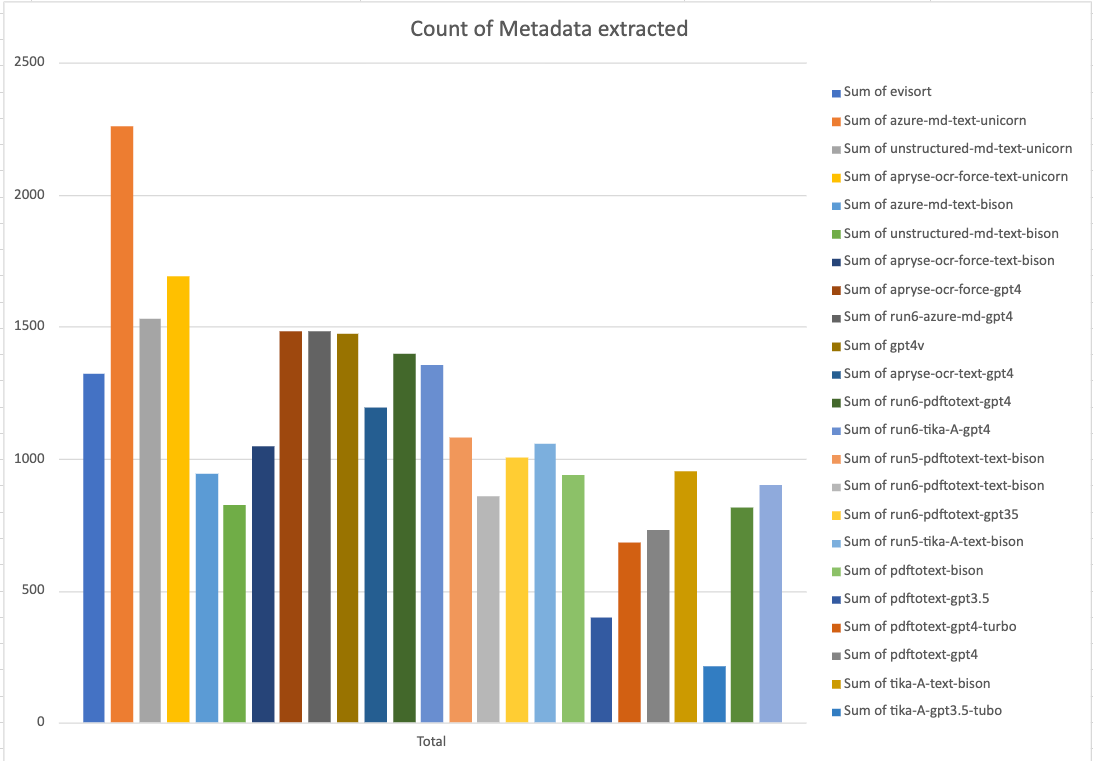}
  \caption{Count of extracted fields}
  \Description{Count of extracted fields for different text conversions}
\end{figure}

Our evaluation of text conversion methods revealed significant variations in performance across different approaches. Using our internal contract dataset, we conducted a systematic comparison of multiple text conversion solutions, including Azure Document Intelligence, Unstructured-IO, Apryse OCR, and others. The comparison maintained consistent LLM configurations across all conversion methods to isolate the impact of text conversion quality.
The results, visualized in Figure 1, demonstrate that Azure Document Intelligence (azure-md) consistently achieves the highest number of successfully extracted fields across different LLM backends. This superior performance, combined with its reasonable operational costs, led to our selection of Azure Document Intelligence as the primary conversion solution in our pipeline.

\subsection{Chunk Selection}
\label{chunk selection exp}

\begin{table}
  \caption{Result on CUAD dataset}
  \label{tab:CUAD}
  \begin{tabular}{cccc}

    Chunk Selection&Context Window&Model Name&F1 Score\\
    \hline
    Oracle & 4096 & Text-Unicorn & 0.94\\

    Baseline & 4096 & Text-Unicorn & 0.66\\

    NER Boosting & 4096 & Text-Unicorn & 0.74\\

    Chunk Reranker & 4096 & Text-Unicorn & 0.76\\

    Baseline & 8192 & Text-Unicorn & 0.75\\

    Chunk Reranker & 8192 & Text-Unicorn & 0.80\\
    \hline
    Baseline & 64000 & Claude 3.5 Sonnet & 0.77\\

    NER Boosting & 8192 & Claude 3.5 Sonnet & 0.80\\
    \hline
    Oracle & 16000 & Gemini 1.5 Flash & 0.90\\

    Chunk Reranker & 16000 & Gemini 1.5 Flash & 0.82\\
    \hline
\end{tabular}
\end{table}

\begin{table}
  \caption{Result on internal lease dataset}
  \label{tab:lease}
  \begin{tabular}{cccc}

    Chunk Selection&Context Window&Model Name&F1 Score\\

    \hline
    Baseline & 16000 & Gemini 1.5 Flash & 0.36\\

    NER Boosting & 16000 & Gemini 1.5 Flash & 0.40\\

    Chunk Reranker & 16000 & Gemini 1.5 Flash & 0.43\\
    \hline
\end{tabular}
\end{table}

To establish a theoretical performance ceiling for chunk selection strategies, we first conducted oracle chunk ranking experiments. In these experiments, we augmented each field's prompt with its corresponding ground truth value, creating what we term the \texttt{oracle\_prompt\_per\_field}. 
The oracle ranking score for each chunk was then computed as the cosine similarity between the embeddings of this augmented prompt and the chunk text. This approach simulates perfect chunk selection by leveraging knowledge of the correct extraction values, providing an upper bound on potential performance.

The oracle experiments achieved remarkable results with an F1 score of 0.94, significantly outperforming current methods. This performance gap motivated the development of our model-based chunk ranking approach.

For training the chunk ranking model, we sampled 10\% and 5\% of documents from the CUAD dataset for training and testing sets respectively. The training data preparation involved several key steps. First, we processed each document through our production chunking pipeline to generate consistent chunk representations. Each data point consisted of chunk text, chunk index, and file name. We then performed a Cartesian product between these chunks and the extraction template fields, followed by joining with ground truth data to establish training targets. The feature set for each chunk-field pair included pre-computed embeddings and their cosine similarities, BM25 similarity scores, and linguistic features derived from Named Entity Recognition (NER) and Part-of-Speech (POS) tagging.

Our model-based approach demonstrated substantial improvements over baseline methods, particularly in constrained context settings. With an 8192 token context window, the model achieved an F1 score of 0.80 compared to baseline scores of 0.75. The improvement was even more pronounced with a 4096 token context window, where the model-based approach achieved an F1 score of 0.76 versus the baseline's 0.66.

To assess generalization capability, we evaluated the system on our internal dataset which is more challenging as it has more than 40 fields to extract at once. We see similar improvement when we switch chunk selection mechanisms with baseline to be 0.36, NER boosing 0.40 and model-based reranker has F1 score 0.43.

The performance gap between the oracle chunk selection and the current chunk re-ranker suggests that while our approach is effective, there remains room for improvement in model generalization. These results indicate that exploring richer feature representations and more sophisticated model architectures could yield further performance gains, potentially closing the gap with oracle performance.

\subsection{LLM configurations}

\subsubsection{Chain of Thought prompting}

As mentioned before, the CoT prompt can significantly improve LLM's ability in reasoning. Here is a part of a concrete example where the LLM tries to extract the end date of a contract in the internal dataset.

\begin{verbatim}
The document does not explicitly state an end date. 

<thinking>

We can deduce that the end date would be
50 days after the effective date.

</thinking>

<thinking>

Calculate the end date by adding 50 days to February 2nd, 2024.  

February 2nd, 2024 + 50 days = March 24th, 2024

</thinking>

<thinking>

We can confidently extract "End Date: March 24, 2024".

</thinking>
\end{verbatim}

Clearly, CoT helped LLM to get the final correct value.
\begin{figure}[h]
  \centering
  \includegraphics[width=\linewidth]{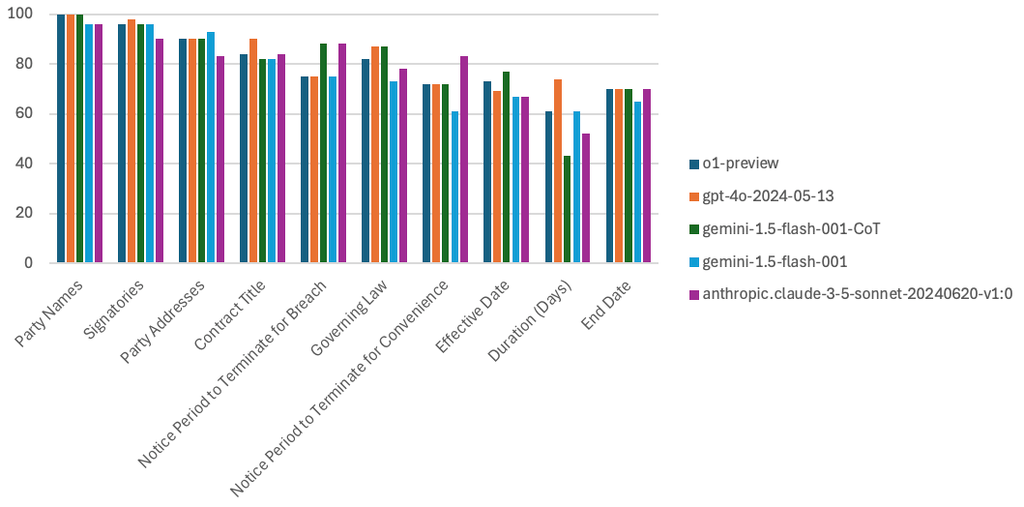}
  \caption{CoT experiment on the internal contract dataset}
  \Description{Cot experiment on the Broadcom datset (internal)}
  \label{fig: cot exp}
\end{figure}

Figure~\ref{fig: cot exp} compares the performance of different LLMs with and without CoT prompting on an internal dataset consisting of 58 contracts. In the baseline comparison without CoT, OpenAI's gpt4o and o1-preview model demonstrate strong performance across most fields, with extraction rates above 90\% for basic fields like Party Names and Signatures. For fields requiring complex reasoning such as Notice Period or Duration calculations, we observe notable variations. When CoT prompting is applied, Gemini Flash shows substantial improvements over its base version, particularly for fields requiring multi-step reasoning. 

These variations reflect the inherent complexity differences in extracting different field types, from simple pattern matching to sophisticated reasoning about contract terms and conditions. Notably, the strong performance of Gemini Flash with CoT prompting has significant practical implications, as it offers substantially lower per-token costs compared to other models. This cost advantage, combined with its competitive extraction accuracy when using CoT prompting, makes it an attractive option for large-scale deployment in production environments where cost efficiency is crucial.

\subsubsection{Structured Output through Tool Calling}

\begin{table}
  \caption{Result on internal lease dataset using json tool}
  \label{tab:lease tool}
  \begin{tabular}{ccc}
    Json Tool&Model Name&F1 Score\\
    \hline
    Yes & Claude 3.5 Sonnet & 0.48\\

    No &  Claude 3.5 Sonnet & 0.39\\
    \hline
    Yes & Gemini 1.5 Flash & 0.36\\

    No &  Gemini 1.5 Flash & 0.36\\
    \hline
    Yes & GPT-4o & 0.42\\

    No &  GPT-4o & 0.40\\
    \hline
\end{tabular}
\end{table}

Tool calling capabilities in modern LLMs provide a mechanism for ensuring structured output format. By defining a JSON schema that specifies the expected structure and type constraints for each metadata field, we leverage the LLM's tool calling interface to enforce consistent output formatting.

To evaluate the impact of JSON tool calling, we conducted experiments on our internal lease dataset, which presents a particularly challenging extraction task with over 40 fields per document. Table ~\ref{tab:lease tool} presents the comparison of extraction performance with and without JSON tool calling across different LLM models.

The results show that JSON tool calling's effectiveness varies across models. Claude 3.5 Sonnet demonstrates the most significant improvement, with F1 score increasing from 0.39 to 0.48 when using the JSON tool. GPT-4o shows a modest improvement from 0.40 to 0.42, while Gemini 1.5 Flash maintains consistent performance at 0.36 regardless of tool usage. The varying impact suggests that the benefit of structured output constraints depends on the model's inherent capabilities in handling complex multi-field extraction tasks.

These findings indicate that while JSON tool calling can enhance extraction accuracy for some models, its effectiveness should be evaluated on a model-specific basis, particularly for challenging datasets with numerous fields. The substantial improvement observed with Claude 3.5 Sonnet suggests that models with strong systematic reasoning capabilities may benefit more from structured output constraints.

\section{LLM as A Judge}

The final component of our system leverages LLMs in an evaluative capacity, both for improving extraction quality and monitoring system performance. We explore two distinct applications: using LLMs to correct extraction errors and implementing online performance monitoring.

\subsection{Label correction}

\begin{table}
  \caption{Match Percent Results on CUAD dataset}
  \label{tab:grader v.s. agent}
  \begin{tabular}{cccc}
    &Grader v.s. Agent&Grader v.s. GT&Agent v.s. GT\\

    All cases & 86.3\% & 80.5\% & 73.3\%\\

    Hard cases & 78.4\% & 76.6\% & 61.1\%\\

\end{tabular}
\end{table}

\begin{table}
  \caption{Extraction with Retries on CUAD dataset}
  \label{tab:retries}
  \begin{tabular}{ccc}
    Number of Retries&Model Name&F1 Score\\

    0 & Text-Unicorn & 0.80\\

    1 & Text-Unicorn & 0.81 \\
    
    0 & Gemini 1.5 Flash & 0.82\\

    1 & Gemini 1.5 Flash & 0.84 \\

\end{tabular}
\end{table}
We developed a specialized LLM-based grading system that evaluates and potentially corrects the outputs of our extraction agent. The grading LLM receives the same input as the extraction agent (including the document text and metadata template) plus the agent's extracted values. It produces two outputs: a numerical score between 0 and 1 for each field and corrected values where it identifies potential errors.

Our experiments revealed an interesting phenomenon (see Table ~\ref{tab:grader v.s. agent}): the grading LLM's extracted values showed higher agreement with ground truth than the original extraction agent's outputs. On comprehensive evaluation, the grading LLM achieved a 80.5\% match rate with ground truth compared to the agent's 73.3\% match rate. This improvement was even more pronounced for challenging cases, where the grading LLM achieved a 76.6\% match rate compared to the agent's 61.1\%.

This performance improvement can be attributed to the grading LLM's advantage of working with pre-extracted candidates, effectively converting the problem from open-ended extraction to validation and correction. The grader's performance suggests that scrutinizing existing candidates may be cognitively easier than identifying relevant information from scratch. This motivates us to implement a retry mechanism when the system fails to extract certain fields value in the first attempt. The already extracted values will be included in the retry and we see clear performance improvement in the retried extraction results (Table ~\ref{tab:retries}).

\subsection{Online grading}

\begin{figure}[h]
  \centering
  \includegraphics[width=\linewidth]{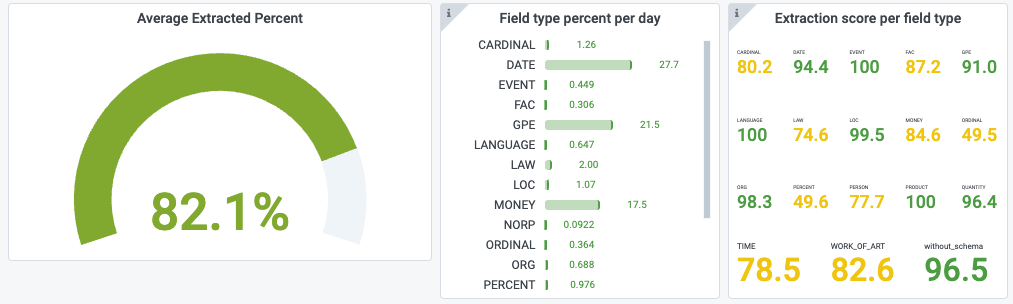}
  \caption{Online monitoring based on LLM grader}
  \Description{online monitoring}
  \label{online eval}
\end{figure}
Due to privacy constraints, we cannot log user prompts or documents for direct performance evaluation. Instead, we leverage the LLM grader for real-time performance monitoring. Our monitoring system, as illustrated in Figure~\ref{online eval}, provides a comprehensive view of system performance through three key dashboards.

The first dashboard shows the overall extraction success rate, currently at 82.1\% across all document types. The second dashboard displays the distribution of field types in extraction requests over time. For instance, during the monitored period, DATE fields constituted 27.7\% of all extraction requests, followed by GPE (Geo-Political Entity) fields at 21.5\%, and MONEY-related fields at 17.5\%. This distribution insight helps us prioritize optimization efforts for frequently requested field types.

The third dashboard presents extraction quality scores broken down by field type. Our system demonstrates particularly strong performance in extracting EVENT (100\%), LANGUAGE (100\%), and PRODUCT (100\%) fields, while showing lower accuracy for ORDINAL (49.5\%) fields. These quality metrics are computed by the LLM grader based on the extraction agent's outputs.

The monitoring framework described above provides crucial insights into system behavior and user patterns while maintaining strict privacy standards. By analyzing aggregate performance metrics and field type distributions, we gain a deep understanding of both system capabilities and user requirements as they evolve over time. These insights drive our optimization efforts and development roadmap, enabling continuous system improvement through anonymous, aggregated data.

All of this is achieved while maintaining strict user data privacy, as the monitoring system operates on aggregate metrics and quality scores rather than raw document content.

\section{Conclusion and Future Works}
In this paper, we presented a comprehensive approach to metadata extraction using Large Language Models. Our work makes several key contributions. First, we identified and optimized three critical components that significantly impact extraction quality: text conversion, chunk selection, and LLM-specific techniques. Through extensive experimentation, we demonstrated that Azure Document Intelligence provides optimal text conversion quality, while our novel NER-enhanced Borda re-ranking method significantly improves chunk selection accuracy. We also showed that Chain of Thought prompting and structured tool calling enhance LLM performance in metadata extraction tasks.

Second, we introduced an innovative LLM-based grading system that serves dual purposes: improving extraction accuracy through label correction and enabling privacy-preserving online performance monitoring. This system achieved a 80.5\% match rate with ground truth, surpassing the base extraction agent's performance of 73.3\%.

Despite these advances, several challenges remain for future research. First, the inherent uncertainty in LLM outputs presents a stability challenge. Even with temperature parameters set to zero, we observe variations between runs that can affect reliability in production environments. This suggests a need for robust uncertainty quantification and mitigation strategies.

Second, while our current chunk selection approach shows promising results, there is potential for developing field-type-specific ranking models. Different metadata fields (e.g., dates, monetary values, legal clauses) may benefit from specialized selection strategies that consider their unique characteristics and common document locations.

Text conversion quality remains a critical area for improvement. We are currently collaborating with a specialized OCR company to develop more advanced and cost-effective text conversion solutions. Their specialized OCR models show promise in handling complex document layouts and improving text extraction quality while reducing operational costs.

Finally, we identified particular challenges in handling multiSelect fields with numerous options. These fields often require complex reasoning across multiple document sections and careful consideration of multiple valid alternatives. Developing specialized techniques for such fields, perhaps incorporating hierarchical reasoning or structured validation steps, represents a promising direction for future research.

As legal AI systems continue to evolve, our work provides a foundation for more accessible, efficient, and accurate contract review tools that could significantly impact legal practice. While these systems are not meant to replace legal professionals, they can serve as valuable tools to augment human expertise and democratize access to contract review services.
\bibliographystyle{ACM-Reference-Format}
\bibliography{main}










\end{document}